\documentclass[11pt, intlimits]{article}
\usepackage[letterpaper, margin=1.0in, footskip=30pt]{geometry}

\usepackage{amsmath}
\usepackage[T1]{fontenc}
\usepackage{mathrsfs}
\usepackage{graphicx}
\usepackage{pgfplots}
\usepackage{float}
\usepackage{tabu}
\usepackage{booktabs}
\usepackage{tikz}
\usepackage{tikz-cd}
\usepackage{url}
\graphicspath{{graphics/}}

\usepackage{caption}
\usepackage[backend=biber, sorting=none, style=numeric-comp]{biblatex}
\begin{document}

\thispagestyle{empty}

\begin{center}
{\LARGE Adversarial Autoencoders in Operator Learning}
\\
25 November 2024
\end{center}

\noindent
\begin{minipage}[t]{0.5\textwidth}
\begin{center}
Dustin Enyeart \\
Department of Mathematics \\
Purdue University \\
\url{denyear@purdue.edu} \\
\end{center}
\end{minipage}%
\hfill
\begin{minipage}[t]{0.5\textwidth}
\begin{center}
Guang Lin\footnotemark \\
Department of Mathematics \\
School of Mechanical Engineering \\
Purdue University \\
\url{guanglin@purdue.edu} 
\end{center}
\end{minipage}
\footnotetext{Corresponding author}

\begin{abstract}
\noindent DeepONets and Koopman autoencoders are two prevalent neural operator architectures. These architectures are autoencoders. An adversarial addition to an autoencoder have improved performance of autoencoders in various areas of machine learning. In this paper, the use an adversarial addition for these two neural operator architectures is studied.  
\end{abstract}

\section{Introduction}
A \emph{neural operator}\index{neural operator} is a neural network that is intended to approximate an operator between function spaces \cite{kovachki2024operator, boulle2023mathematical, winovich2021neural}. 
An example of an output function for a neural operator is a solution to a differential equation. 
Examples of input functions for a neural operator are the initial conditions or boundary conditions for the differential equation. 

DeepONets are Koopman autoencoders are both neural operator architectures.
Furthermore, they are both autoencoders, that is, they consist of an encoder to a latent space and a decoder from this latent space. 
An adversarial addition to autoencoders is a common extension for neural networks. 
Intuitively, an adversarial addition to an autoencoder is to encourage the encoder to use the entire latent space in a continuous way, that is, near inputs are near in the latent space. 

First, in this paper, autoencoders and adversarial autoencoders are explained. 
Then, the neural operator architectures DeepONets and Koopman autoencoders are introduced.
Then, the differential equations that are used in the numerical experiments are presented. 
Then, results from numerical experiments are presented.
Finally, the paper ends with a conclusion of the numerical experiments. 

\section{Adversarial Autoencoders}
\emph{Autoencoders}\index{autoencoder}\index{autoencoder} are a neural network architecture that consists of the composition of an encoder and a decoder \cite{hinton2006reducing, rumelhart1986learning}. 
The \emph{encoder}\index{encoder} is a neural network that maps the input into a \emph{latent space}\index{latent spcae}, and the \emph{decoder}\index{decoder} is a neural network from the latent space to the original space.  
The input and output are in the same space, and the loss of an autoencoder is a measure of the difference between the input and output.
Normally, for neural networks, the latent space is smaller than the input space, and the autoencoder tries to learn a lower dimensional representation, that is, it compresses the input\index{data compression}\index{manifold learning}\index{semisupervised learning}. 

\emph{Adversarial Autoencoders}\index{adversarial autoencoder} are an extension of autoencoders to include a discriminator \cite{kingma2013auto, makhzani2015adversarial}.
The \emph{discriminator}\index{discriminator} is a neural network from the latent space to the real numbers between $0$ and $1$. 
The discriminator learns to distinguish whether a point in the latent space is in the image of the encoder. 
Furthermore, the encoder also tries to learn to fool the discriminator. 
The idea is that this encourages the encoder to use the entire latent space. 
Data for points in the range of the encoder is from encoding points from the input space, and data that is not in the range of the encoder is from picking random points in the latent space. 
Such random points are picked from a probability distribution that is similar to the distribution of the true encodings, such as the normal distribution with the same mean and standard deviation as the true encodings.

\begin{figure}[H]
\[
\begin{tikzcd}
    x \arrow{r}{\mathrm{encoder}} & \arrow{r}{\mathrm{decoder}} & y 
    \\
    \\
\end{tikzcd}
\hspace{4cm}
\begin{tikzcd}
    x \arrow{r}{\mathrm{encoder}}
    & \arrow{r}{\mathrm{decoder}} \arrow{d}{\mathrm{discriminator}}
    & y 
    \\
    & z & 
    \\
\end{tikzcd}
\]
\caption{An autoencoder and an adversarial autoencoder: The left is an autoencoder, where an encoder maps the input $x$ into a latent space, and a decoder maps this encoding to the output $y$. The right is an adversarial autoencoder, which extends the autoencoder to include a discriminator that maps an encoding to an output $z$. 
}
\end{figure}
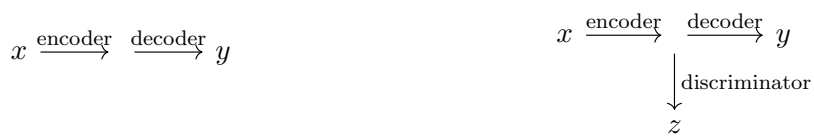

The discriminator is trained separately from the encoder-decoder pair so that they oppose each other instead of help each other. 
In practice, they are trained on alternative epochs. 
Thus, the encoder-decoder pair improves at fooling the discriminator, then the discriminator improves at distinguishing what came from the input, and the process repeats. 

Intuitively, the output of the discriminator is a probability. 
This is normally done by using the sigmoid function as the activation function. 
The \emph{sigmoid function}\index{sigmoid function} is the function that maps an input $x$ to 
\[
    \frac{1}{1 + \exp(-x)}.
\]
Furthermore, the loss for the discriminator is computed with the binary cross entropy loss. 
The \emph{binary cross entropy loss}\index{binary cross entropy loss} between a prediction $a$ and a target $b$ is
\[
    b \cdot \log(a) + (1 - b) \cdot \log(1 - a)
    .
\]
It is popular for binary classification tasks\index{classification} \cite{ruby2020binary}.
When the encoder-decoder pair is being trained, the binary cross entropy loss is used with the opposite targets. 
For the experiments in this chapter, this loss term is weighted by the accuracy loss so that this term does not dominate the training.

\begin{figure}[H]
\begin{center}
    \includegraphics[scale=.6]{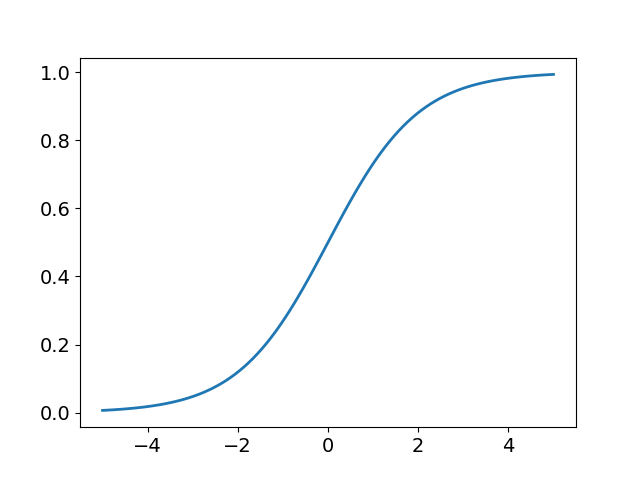}
    \caption{The sigmoid function
        }
\end{center}
\end{figure}

Adversarial autoencoders are particularly popular for generative models\index{generative learning} \cite{foster2019generative}. 
The idea is that once the encoder is trained to fool the discriminator, a decoding of a random point in the latent space will share characteristics similar to the input data. 
For example, if the autoencoder learns to compress images of a certain style, then the decoding of a random point in the latent space will hopefully be an image of that style.

In an adversarial autoencoder, it is common to add noise\index{noise} to the output of the encoder before passing it to the discriminator.
This helps the model be more robust instead of memorizing the encodings of the input data and encourages continuity. 
This can be thought of as an aid to having only a finite amount of data. 

\section{Architectures}
This section introduces the architectures used in this paper. 
These are DeepONets and Koopman autoencoders.

\subsection{DeepONets}

\emph{Deep neural operators}\index{deep neural operator}, which are abbreviated as \emph{DeepONets}\index{DeepONet}, are a neural operator architecture \cite{DeepONet, lanthaler2022error, goswami2022physics, he2023novel, cho2024learning, lu2022comprehensive}. 
They consist of two parts. 
The first part encodes information about the differential equation into a latent space.
It is called the \emph{branch network}\index{branch network}.
The second part encodes a position into a latent space.
It is called the \emph{trunk network}\index{trunk network}.
An example input for the branch network is the initial condition for a dynamic differential equation, and an example for the trunk network is the space-time point to evaluate the solution to the differential equation.
In practice, there is usually only one forward pass for the branch network, but many forward passes for the trunk network to encode all the desired positions.

The dimensions of the latent spaces of the branch network and the trunk network are the same. 
And, the output of such a model is the dot product between both latent spaces. 
Intuitively, the branch network encodes the solution to the differential equation as a set of basis functions, and the trunk network encodes the position as the coefficients of these basis functions.

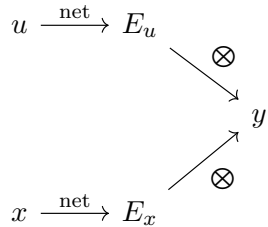
\begin{figure}[H]
\[
\begin{tikzcd}
    u \arrow{r}{\mathrm{net}} & E_{u} \arrow{dr}{\bigotimes} & 
    \\
    & & y
    \\
    x \arrow{r}{\mathrm{net}} & E_{x} \arrow[swap]{ur}{\bigotimes} & 
    \\
\end{tikzcd}
\]
\caption{The DeepONet architecture:
    The input $u$ is the input function, and the input $x$ is the point where the output function is evaluated. Their encodings are denoted by $E_u$ and $E_x$, respectively. 
    The output is denoted by $y$.}
\end{figure}

\subsection{Koopman Autoencoders}

\emph{Koopman autoencoders}\index{Koopman autoencoder} are a neural operator architecture that are used for time-dependent differential equations \cite{lusch2018deep, koop1, mamakoukas2020learning, huang2020data, klus2020data}. 
They are popular\index{applications of Koopman autoencoders} for dynamic mode decomposition\index{dynamic mode decomposition} \cite{dymodedecomp1, dymodedecomp2, kutz2016multiresolution, takeishi2017learning, bagheri2013koopman} and control\index{control} \cite{budivsic2020koopman, li2019learning, kaiser2020data, han2020deep, bruder2021koopman, bruder2019modeling, arbabi2018data, abraham2017model}. 

The Koopman formulation of classical mechanics\index{Koopman formulation of mechanics} is an alternative framework of classical mechanics \cite{koopman1931hamiltonian, brunton2021modern, bruce2019koopman}. 
It was inspired by the Hamiltonian formulation of quantum mechanics \cite{morrison1990understanding, ohanian1989principles, griffiths2018introduction}. 
In this theory, the physical state\index{physical space} at a given time is represented as a state in an infinite-dimensional latent space\index{latent space}, and an infinite-dimensional operator governs the time evolution in this latent space.
Intuitively, the Koopman formulation can be thought of as removing nonlinearity by infinitely increasing the dimension. 

The Koopman formulation of classical mechanics can be discretized\index{discretized Koopman formulation} to provide a numerical scheme by approximating the time-evolution operator by a finite-dimensional matrix $K$.
In this scheme, a physical state $s_0$ can be evolved into a later physical state $s_n$ by encoding it into the latent space, applying the matrix $K$ repetitively and then decoding it back to the physical space, that is, the equation 
\[
    s_n = R \circ K^{n} \circ E(s_0)
\]
approximately holds, where $E$ is a discretized encoder and $R$ is a discretized decoder.
In this numerical scheme, the dimension of this latent space is called the \emph{encoding dimension}\index{encoding dimension}.

\begin{figure}[H]
\[
\begin{tikzcd}
    e_0 \arrow{r}{K}
    & e_1 \arrow{d}{R} \arrow{r}{K}
    & e_2 \arrow{d}{R} \arrow{r}{K}
    & \cdots \arrow{r}{K}
    & e_n \arrow{d}{R} \\
    s_0 \arrow[swap]{u}{E} \arrow{r}{f}
    & s_1 \arrow{r}{f}
    & s_2 \arrow{r}{f}
    & \cdots \arrow{r}{f}
    & s_n \\
\end{tikzcd}
\]
\caption{Discretization of the Koopman formulation into a numerical scheme: The physical states at successive time points are denoted by $s_0$, $s_1$, $\dots$, $s_{n-1}$ and $s_n$. The encoded states at successive time points are denoted by $e_0$, $e_1$, $\dots$, $e_{n-1}$ and $e_n$. The function $f$ is the true time evolution of the physical state by the time step. The discretized Koopman operator, encoder and decoder are denoted by $K$, $E$ and $R$, respectively.}
\end{figure}    

\emph{Koopman autoencoders}\index{Koopman autoencoder}\index{Koopman architecture}\index{Koopman neural network} are based on the discretized Koopman formulation where the operator\index{Koopman operator}, encoder\index{Koopman encoder} and decoder\index{Koopman decoder} are neural networks.
The encoder and decoder normally consists of fully connected layers or convolution layers. 
And, the operator is a fully connected layer without a bias, and an activation function\index{activation function} is not used between successive applications of it.  
Because the operator is applied repeatedly, it can be useful to use gradient clipping. 

The loss for DeepONets and Fourier neural operators is the mean squared error between the output of the model and the true solution.
For Koopman autoencoders, it is better to use additional loss terms during training \cite{enyeart2024loss}. 
The training loss used for Koopman autoencoders in this paper is 
\[
\begin{split}
    & \frac{1}{n} \cdot \sum_i \vert \vert R \circ K^i \circ E(v_0) - v_i \vert \vert^2 \\
    & + \frac{1}{n} \cdot \sum_{i} \vert \vert R \circ E(v_i) - v_i \vert \vert^2 \\
    & + \vert \vert K \circ K^{\mathrm{T}} - I \vert \vert^2, 
\end{split}
\]
where $E$ is the encoder, $R$ is the encoder, $K$ is the Koopman operator, $n$ is the number of time steps and each $v_i$ is a vector in the physical space. 
Intuitively, the first term is the accuracy, the second term regulates the relation between the encoder and the decoder, and the last term forces the operator to be unitary. 
Furthermore, a mask is on the Koopman operator so that it is tridiagonal. 
The intuition of these choices is from the Koopman formulation of classical mechanics. 

\section{Differential Equations}
This section introduces the differential equations that are used for the numerical experiments in this paper.
The ordinary differential equations are the equation for the pendulum, the Lorenz system and a fluid attractor equation. 
The partial differential equations are Burger's equation and the Korteweg-de-Vries equation.

\subsection{Pendulum}

The equation for the \emph{pendulum}\index{pendulum} is the differential equation
\[
    \frac{\mathrm{d}^2 \theta}{\mathrm{d}^2 t}
    = - \sin (\theta)
    . 
\]
This is a time-dependent second-order nonlinear ordinary differential equation whose dimension is $1$. 
This equation models the motion of a pendulum in a constant gravitational field, where the variable $\theta$ is the angle of the pendulum from vertical.

In the numerical experiments in this paper, the models attempt to learn the solution to this differential equation as a function of the initial condition.
To get data, initial positions are generated randomly.
Then, this differential equation is numerically solved for these initial conditions using the Runge-Kutta Method \cite{leveque2007finite}.

\subsection{Lorenz System}

The \emph{Lorenz system}\index{Lorenz system} is the differential equation
\[
\begin{cases}
    \frac{\mathrm{dx}}{\mathrm{d}t} = y - x  \\
    \frac{\mathrm{dy}}{\mathrm{d}t} = x - x \cdot z - y  \\
    \frac{\mathrm{dz}}{\mathrm{d}t} = x \cdot y - z. \\ 
\end{cases}
\]
This is a time-dependent first-order nonlinear ordinary differential equation whose dimension is $3$. 
Historically, this equation was used in weather modeling.
Now, it provides a benchmark example of a chaotic system\index{chaotic system}.

In the numerical experiments in this paper, the models attempt to learn the solution to this differential equation as a function of the initial condition.
To get data, initial positions are generated randomly.
Then, this differential equation is numerically solved for for these initial conditions using the Runge-Kutta Method \cite{leveque2007finite}.

\subsection{Fluid Attractor Equation}

The differential equation 
\[
\begin{cases}
    \frac{\mathrm{dx}}{\mathrm{d}t} = x - y + x \cdot z  \\
    \frac{\mathrm{dy}}{\mathrm{d}t} = x + y + y \cdot z \\
    \frac{\mathrm{dz}}{\mathrm{d}t} = x^2 + y^2 + z \\ 
\end{cases}
\]
is used to model fluid flow around a cylinder\index{fluid attractor equation} \cite{noack2003hierarchy}. 
It is a time-dependent first-order nonlinear ordinary differential equation whose dimension is $3$.

In the numerical experiments in this paper, the models attempt to learn the solution to this differential equation as a function of the initial condition.
To get data, initial positions are generated randomly.
Then, this differential equation is numerically solved for for these initial conditions using the Runge-Kutta Method \cite{leveque2007finite}.

\subsection{Burger's Equation}

\emph{Burger's equation}\index{Burger's equation} is the differential equation
\[
    \frac{\partial u}{\partial t}
    = \frac{\partial^2 u}{\partial^2 x} - u\frac{\partial u}{\partial x} 
    .
\]
This is a time-dependent first-order nonlinear partial differential equation whose domain dimension and range dimension are both $1$. 
It is used to model some fluids. 

In the numerical experiments in this paper, the models attempt to learn the solution to this differential equation as a function of the initial condition.
Furthermore, Dirichlet boundary conditions are used for this equation such that the value of the unknown function is $0$ on the boundary.
To get data, initial conditions are made by generating random symbolic expressions that satisfy the boundary conditions, and these expressions are then numerically sampled. 
Then, the partial differential equation is numerically solved for these initial conditions \cite{leveque1992numerical}.

\subsection{Korteweg-de-Vries Equation}

The \emph{Korteweg-de-Vries equation}\index{Korteweg-de-Vries equation}, which is abbreviated as the \emph{KdV equation}\index{KdV equation}, is the differential equation
\[
    \frac{\partial u}{\partial t}
    = 6u\frac{\partial u}{\partial x}
    - \frac{\partial^3u}{\partial ^3x}
    .
\]
This is a time-dependent first-order nonlinear partial differential equation whose domain dimension and range dimension are both $1$. 
It is used to model some fluids. 

In the numerical experiments in this paper, the models attempt to learn the solution to this differential equation as a function of the initial condition.
Furthermore, periodic boundary conditions are used for this equation, that is, the value of the unknown function is the same on each endpoint of the spatial domain. 
To get data, initial conditions are made by generating random symbolic expressions that satisfy the boundary conditions, and these expressions are then numerically sampled. 
Then, the partial differential equation is numerically solved for these initial conditions \cite{zabusky1965interaction}.

\section{Numerical Experiments}
Five numerical experiments are are presented. 
Two experiments are done for DeepONets, which are for Burger's equation and the KdV equation. 
Three experiments are done for Koopman autoencoders, which are for the pendulum, the Lorenz system, and the fluid attractor equation.
The results are presented in Table \ref{system-deeponet} and Table \ref{system-koopman}, respectively. 

Noise was added to the output of the encoder before passing it to the discriminator. 
The noise was selected from the normal distribution whose mean is $0$ and whose standard deviation is the standard deviation of the true encodings scaled by $0.025$.
Furthermore, the models were trained using stochastic weight averaging \cite{enyeart2024best}.

\begin{table}[H]
    \caption{Results for DeepONets: The left table is for Burger's equation. The right table is for the KdV equation.}
    \label{system-deeponet}
    \begin{minipage}{.5\textwidth}
        \vspace{1em}
        \begin{center}
        Burger's \\
        \begin{tabular}{lr}
            error & adversarial \\
            \toprule
            3.860e-2 & false \\
            3.707e-2 & true \\
        \end{tabular}
        \end{center}
    \end{minipage}
    \hspace{1cm}
    \begin{minipage}{.5\textwidth}
        \vspace{1em}
        \begin{center}
        KdV \\
        \begin{tabular}{lr}
            error & adversarial \\
            \toprule
            3.059e-2 & false \\
            2.795e-2 & true \\
        \end{tabular}
        \end{center}
    \end{minipage}
\end{table}

\begin{table}[H]
    \caption{Results for Koopman autoencoders: The left table is for the equation for the pendulum. The center table is for the Lorenz system. The right table is for the fluid attractor equation.}
    \label{system-koopman}
    \begin{minipage}{.32\textwidth}
        \vspace{1em}
        \begin{center}
        pendulum \\
        \begin{tabular}{lr}
            error & adversarial \\
            \toprule
            2.615e-3 & false \\
            1.922e-3 & true \\
        \end{tabular}
        \end{center}
    \end{minipage}
    \begin{minipage}{.32\textwidth}
        \vspace{1em}
        \begin{center}
        Lorenz \\
        \begin{tabular}{lr}
            error & adversarial \\
            \toprule
            5.874e-2 & false \\
            4.703e-2 & true \\
        \end{tabular}
        \end{center}
    \end{minipage}
    \begin{minipage}{.32\textwidth}
        \vspace{1em}
        \begin{center}
        fluid attractor \\
        \begin{tabular}{lr}
            error & adversarial \\
            \toprule
            2.833e-5 & false \\
            2.637e-5 & true \\
        \end{tabular}
        \end{center}
    \end{minipage}
\end{table}

The authors did numerical experiments with various amounts of training data. 
With a large amount of data, the adversarial addition did not help and would normally preform slightly less. 
The experiments presented here are for a small amount of training data. 
The number of training samples used were $25$ for Burger's equation, $50$ for the KdV equation, $20$ for the pendulum, $48$ for the Lorenz system, and $40$ for the fluid attractor.
This roughly the minimal amount of data needed for these models to learn to a reasonable accuracy.

\begin{figure}[H]
\begin{center}
    \includegraphics[scale=.5]{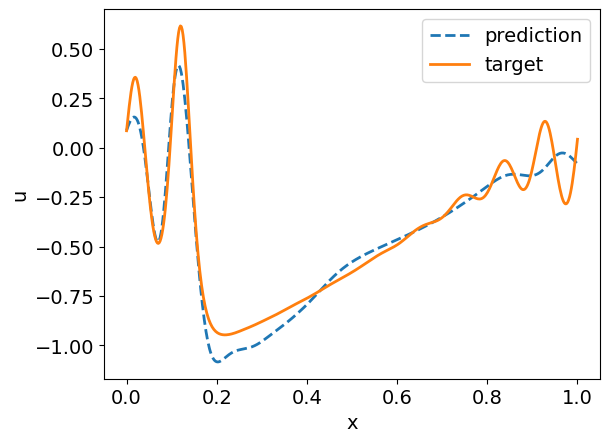}
    \includegraphics[scale=.5]{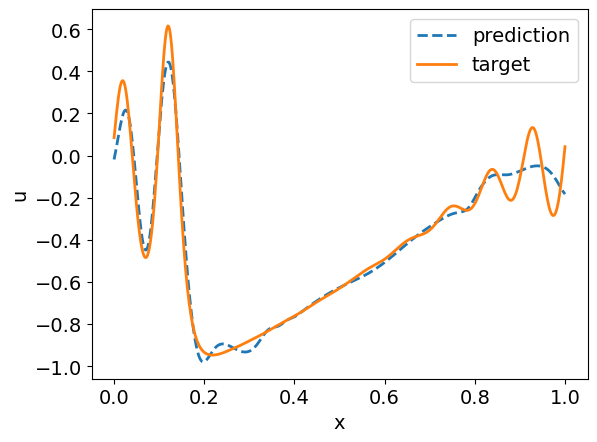}
    \caption{Comparison of an adversarial addition to a DeepONet for the KdV equation: Both plots are of the final time slices of the target and prediction. The left plot is without an adversarial addition. The right plot is with an adversarial addition. The model with the adversarial addition is visibly more accurate, and it particularly hugs the true solution more at the main bend. Both models are still relatively inaccurate compared to models trained with a large amount of data.}
\end{center}
\end{figure}

For DeepOnets, the performance improvements for using an adversarial addition are $4.0\%$ for Burger's equation and $8.6\%$ for the KdV equation. 
For Koopman autoencoders, the performance improvements are $26.5\%$ for the pendulum, $19.9\%$ for the Lorenz system, and $6.9\%$ for the fluid attractor.

\section{Conclusion}
The adversarial addition routinely improved the accuracy of the models, both DeepONets an Koopman autoencoders, by about $5\%$ to $10\%$. 
These improvements are only for when a small amount of training data is available. 
Including an adversarial addition increases the expense of training, but, because it is used for small amount of training data, the training is already relatively cheap. 
Thus, it is recommended to use an adversarial addition for these models when only a small amount of training data is available.     

\section*{Code Availability}
The code for the numerical experiments is publicly available at \url{https://gitlab.com/dustin_lee/neural-operators}.
All experiments were done in Python\index{Python}. 
Neural networks were implemented with PyTorch\index{PyTorch} \cite{paszke2017automatic, pointer2019programming, lippe2024uvadlc} and Lightning\index{Lightning} \cite{lightning}. 
Hyperparameter configuration was done using Hydra\index{Hydra} \cite{Yadan2019Hydra}.

\section*{Declaration of Competing Interest}
The authors declare that they have no known competing financial interests or personal relationships that could have appeared to influence the work reported in this paper.

\section*{Acknowledgment}
GL would like to thank the support of National Science Foundation (DMS-2053746, DMS-2134209, ECCS-2328241, CBET-2347401 and OAC-2311848), and U.S.~Department of Energy (DOE) Office of Science Advanced Scientific Computing Research program DE-SC0023161, and DOE–Fusion Energy Science, under grant number: DE-SC0024583.

\printbibliography

@misc{enyeart2024loss,
      title={Loss Terms and Operator Forms of Koopman Autoencoders}, 
      author={Dustin Enyeart and Guang Lin},
      year={2024},
      eprint={2412.04578},
      archivePrefix={arXiv},
      primaryClass={cs.LG},
      url={https://arxiv.org/abs/2412.04578}, 
}

@misc{enyeart2024best,
      title={Some Best Practices in Operator Learning}, 
      author={Dustin Enyeart and Guang Lin},
      year={2024},
      eprint={2412.06686},
      archivePrefix={arXiv},
      primaryClass={cs.LG},
      url={https://arxiv.org/abs/2412.06686}, 
}

@article{rumelhart1986learning,
  title={Learning internal representations by error propagation, parallel distributed processing, explorations in the microstructure of cognition, ed. de rumelhart and j. mcclelland. vol. 1. 1986},
  author={Rumelhart, David E and Hinton, Geoffrey E and Williams, Ronald J},
  journal={Biometrika},
  volume={71},
  number={599-607},
  pages={6},
  year={1986}
}

@article{hinton2006reducing,
  title={Reducing the dimensionality of data with neural networks},
  author={Hinton, Geoffrey E and Salakhutdinov, Ruslan R},
  journal={science},
  volume={313},
  number={5786},
  pages={504--507},
  year={2006},
  publisher={American Association for the Advancement of Science}
}

@article{kingma2013auto,
  title={Auto-encoding variational bayes},
  author={Kingma, Diederik P},
  journal={arXiv preprint arXiv:1312.6114},
  year={2013}
}

@article{makhzani2015adversarial,
  title={Adversarial autoencoders},
  author={Makhzani, Alireza and Shlens, Jonathon and Jaitly, Navdeep and Goodfellow, Ian and Frey, Brendan},
  journal={arXiv preprint arXiv:1511.05644},
  year={2015}
}

@article{foster2019generative,
  title={Generative deep learning: teaching machines to paint},
  author={Foster, David},
  journal={Write, Compose, and Play (Japanese Version) O’Reilly Media Incorporated},
  pages={139--140},
  year={2019}
}

@article{ruby2020binary,
  title={Binary cross entropy with deep learning technique for image classification},
  author={Ruby, Usha and Yendapalli, Vamsidhar},
  journal={Int. J. Adv. Trends Comput. Sci. Eng},
  volume={9},
  number={10},
  year={2020}
}

@article{kovachki2024operator,
  title={Operator learning: Algorithms and analysis},
  author={Kovachki, Nikola B and Lanthaler, Samuel and Stuart, Andrew M},
  journal={arXiv preprint arXiv:2402.15715},
  year={2024}
}

@article{boulle2023mathematical,
  title={A mathematical guide to operator learning},
  author={Boull{\'e}, Nicolas and Townsend, Alex},
  journal={arXiv preprint arXiv:2312.14688},
  year={2023}
}

@phdthesis{winovich2021neural,
  title={Neural Network Approximations to Solution Operators for Partial Differential Equations},
  author={Winovich, Nick},
  year={2021},
  school={Purdue University}
}

@article{koopman1931hamiltonian,
  title={Hamiltonian systems and transformation in Hilbert space},
  author={Koopman, Bernard O},
  journal={Proceedings of the National Academy of Sciences},
  volume={17},
  number={5},
  pages={315--318},
  year={1931},
  publisher={National Acad Sciences}
}

@article{brunton2021modern,
  title={Modern Koopman theory for dynamical systems},
  author={Brunton, Steven L and Budi{\v{s}}i{\'c}, Marko and Kaiser, Eurika and Kutz, J Nathan},
  journal={arXiv preprint arXiv:2102.12086},
  year={2021}
}

@inproceedings{bruce2019koopman,
  title={What is the Koopman operator? a simplified treatment for discrete-time systems},
  author={Bruce, Adam L and Zeidan, Vera M and Bernstein, Dennis S},
  booktitle={2019 American Control Conference (ACC)},
  pages={1912--1917},
  year={2019},
  organization={IEEE}
}

@book{morrison1990understanding,
  title={Understanding quantum physics: A user's manual},
  author={Morrison, Michael A},
  year={1990}
}

@book{ohanian1989principles,
  title={Principles of Quantum Mechanics},
  author={Hans C. Ohanian},
  year={1989}
}

@book{griffiths2018introduction,
  title={Introduction to Quantum Mechanics},
  author={David J. Griffiths},
  year={2018}
}

@article{lusch2018deep,
  title={Deep learning for universal linear embeddings of nonlinear dynamics},
  author={Lusch, Bethany and Kutz, J Nathan and Brunton, Steven L},
  journal={Nature communications},
  volume={9},
  number={1},
  pages={4950},
  year={2018},
  publisher={Nature Publishing Group UK London}
}

@article{koop1,
  title={A survey on the methods and results of data-driven koopman analysis in the visualization of dynamical systems},
  author={Parmar, Nishaal and Refai, Hazem H and Runolfsson, Thordur},
  journal={IEEE Transactions on Big Data},
  volume={8},
  number={3},
  pages={723--738},
  year={2020},
  publisher={IEEE}
}

@article{mamakoukas2020learning,
  title={Learning data-driven stable Koopman operators},
  author={Mamakoukas, Giorgos and Abraham, Ian and Murphey, Todd D},
  journal={Free radical biology \& medicine.},
  year={2020},
  publisher={Elsevier Inc.}
}

@article{huang2020data,
  title={Data-driven nonlinear stabilization using koopman operator},
  author={Huang, Bowen and Ma, Xu and Vaidya, Umesh},
  journal={The Koopman Operator in Systems and Control: Concepts, Methodologies, and Applications},
  pages={313--334},
  year={2020},
  publisher={Springer}
}

@article{klus2020data,
  title={Data-driven approximation of the Koopman generator: Model reduction, system identification, and control},
  author={Klus, Stefan and N{\"u}ske, Feliks and Peitz, Sebastian and Niemann, Jan-Hendrik and Clementi, Cecilia and Sch{\"u}tte, Christof},
  journal={Physica D: Nonlinear Phenomena},
  volume={406},
  pages={132416},
  year={2020},
  publisher={Elsevier}
}

@article{dymodedecomp1,
  title={Koopman invariant subspaces and finite linear representations of nonlinear dynamical systems for control},
  author={Brunton, Steven L and Brunton, Bingni W and Proctor, Joshua L and Kutz, J Nathan},
  journal={PloS one},
  volume={11},
  number={2},
  pages={e0150171},
  year={2016},
  publisher={Public Library of Science San Francisco, CA USA}
}

@article{dymodedecomp2,
  title={Applied koopmanism},
  author={Budi{\v{s}}i{\'c}, Marko and Mohr, Ryan and Mezi{\'c}, Igor},
  journal={Chaos: An Interdisciplinary Journal of Nonlinear Science},
  volume={22},
  number={4},
  year={2012},
  publisher={AIP Publishing}
}

@article{kutz2016multiresolution,
  title={Multiresolution dynamic mode decomposition},
  author={Kutz, J Nathan and Fu, Xing and Brunton, Steven L},
  journal={SIAM Journal on Applied Dynamical Systems},
  volume={15},
  number={2},
  pages={713--735},
  year={2016},
  publisher={SIAM}
}

@article{takeishi2017learning,
  title={Learning Koopman invariant subspaces for dynamic mode decomposition},
  author={Takeishi, Naoya and Kawahara, Yoshinobu and Yairi, Takehisa},
  journal={Advances in neural information processing systems},
  volume={30},
  year={2017}
}

@article{bagheri2013koopman,
  title={Koopman-mode decomposition of the cylinder wake},
  author={Bagheri, Shervin},
  journal={Journal of Fluid Mechanics},
  volume={726},
  pages={596--623},
  year={2013},
  publisher={Cambridge University Press}
}

@misc{budivsic2020koopman,
  title={The Koopman Operator in Systems and Control: Concepts, Methodologies, and Applications},
  author={Budi{\v{s}}ic, M and Mohr, R and Mezic, I},
  year={2020},
  publisher={Springer}
}

@article{li2019learning,
  title={Learning compositional koopman operators for model-based control},
  author={Li, Yunzhu and He, Hao and Wu, Jiajun and Katabi, Dina and Torralba, Antonio},
  journal={arXiv preprint arXiv:1910.08264},
  year={2019}
}

@article{kaiser2020data,
  title={Data-driven approximations of dynamical systems operators for control},
  author={Kaiser, Eurika and Kutz, J Nathan and Brunton, Steven L},
  journal={The Koopman Operator in Systems and Control: Concepts, Methodologies, and Applications},
  pages={197--234},
  year={2020},
  publisher={Springer}
}

@inproceedings{han2020deep,
  title={Deep learning of Koopman representation for control},
  author={Han, Yiqiang and Hao, Wenjian and Vaidya, Umesh},
  booktitle={2020 59th IEEE Conference on Decision and Control (CDC)},
  pages={1890--1895},
  year={2020},
  organization={IEEE}
}

@article{bruder2021koopman,
  title={Koopman-based control of a soft continuum manipulator under variable loading conditions},
  author={Bruder, Daniel and Fu, Xun and Gillespie, R Brent and Remy, C David and Vasudevan, Ram},
  journal={IEEE robotics and automation letters},
  volume={6},
  number={4},
  pages={6852--6859},
  year={2021},
  publisher={IEEE}
}

@article{bruder2019modeling,
  title={Modeling and control of soft robots using the koopman operator and model predictive control},
  author={Bruder, Daniel and Gillespie, Brent and Remy, C David and Vasudevan, Ram},
  journal={arXiv preprint arXiv:1902.02827},
  year={2019}
}

@inproceedings{arbabi2018data,
  title={A data-driven koopman model predictive control framework for nonlinear partial differential equations},
  author={Arbabi, Hassan and Korda, Milan and Mezi{\'c}, Igor},
  booktitle={2018 IEEE Conference on Decision and Control (CDC)},
  pages={6409--6414},
  year={2018},
  organization={IEEE}
}

@article{abraham2017model,
  title={Model-based control using Koopman operators},
  author={Abraham, Ian and De La Torre, Gerardo and Murphey, Todd D},
  journal={arXiv preprint arXiv:1709.01568},
  year={2017}
}

@article{DeepONet,
   title={Learning nonlinear operators via DeepONet based on the universal approximation theorem of operators},
   volume={3},
   ISSN={2522-5839},
   url={http://dx.doi.org/10.1038/s42256-021-00302-5},
   DOI={10.1038/s42256-021-00302-5},
   number={3},
   journal={Nature Machine Intelligence},
   publisher={Springer Science and Business Media LLC},
   author={Lu, Lu and Jin, Pengzhan and Pang, Guofei and Zhang, Zhongqiang and Karniadakis, George Em},
   year={2021},
   month=mar, pages={218–229}
}

@article{lanthaler2022error,
  title={Error estimates for deeponets: A deep learning framework in infinite dimensions},
  author={Lanthaler, Samuel and Mishra, Siddhartha and Karniadakis, George E},
  journal={Transactions of Mathematics and Its Applications},
  volume={6},
  number={1},
  pages={tnac001},
  year={2022},
  publisher={Oxford University Press}
}

@article{goswami2022physics,
  title={A physics-informed variational DeepONet for predicting crack path in quasi-brittle materials},
  author={Goswami, Somdatta and Yin, Minglang and Yu, Yue and Karniadakis, George Em},
  journal={Computer Methods in Applied Mechanics and Engineering},
  volume={391},
  pages={114587},
  year={2022},
  publisher={Elsevier}
}

@article{he2023novel,
  title={Novel DeepONet architecture to predict stresses in elastoplastic structures with variable complex geometries and loads},
  author={He, Junyan and Koric, Seid and Kushwaha, Shashank and Park, Jaewan and Abueidda, Diab and Jasiuk, Iwona},
  journal={Computer Methods in Applied Mechanics and Engineering},
  volume={415},
  pages={116277},
  year={2023},
  publisher={Elsevier}
}

@article{cho2024learning,
  title={Learning time-dependent PDE via graph neural networks and deep operator network for robust accuracy on irregular grids},
  author={Cho, Sung Woong and Lee, Jae Yong and Hwang, Hyung Ju},
  journal={arXiv preprint arXiv:2402.08187},
  year={2024}
}

@article{lu2022comprehensive,
  title={A comprehensive and fair comparison of two neural operators (with practical extensions) based on fair data},
  author={Lu, Lu and Meng, Xuhui and Cai, Shengze and Mao, Zhiping and Goswami, Somdatta and Zhang, Zhongqiang and Karniadakis, George Em},
  journal={Computer Methods in Applied Mechanics and Engineering},
  volume={393},
  pages={114778},
  year={2022},
  publisher={Elsevier}
}

@book{leveque2007finite,
  title={Finite difference methods for ordinary and partial differential equations: steady-state and time-dependent problems},
  author={LeVeque, Randall J},
  year={2007},
  publisher={SIAM}
}

@book{leveque1992numerical,
  title={Numerical methods for conservation laws},
  author={LeVeque, Randall J and Leveque, Randall J},
  volume={214},
  year={1992},
  publisher={Springer}
}

@article{noack2003hierarchy,
  title={A hierarchy of low-dimensional models for the transient and post-transient cylinder wake},
  author={Noack, Bernd R and Afanasiev, Konstantin and Morzy{\'n}ski, Marek and Tadmor, Gilead and Thiele, Frank},
  journal={Journal of Fluid Mechanics},
  volume={497},
  pages={335--363},
  year={2003},
  publisher={Cambridge University Press}
}

@article{zabusky1965interaction,
  title={Interaction of ``solitons" in a collisionless plasma and the recurrence of initial states},
  author={Zabusky, Norman J and Kruskal, Martin D},
  journal={Physical review letters},
  volume={15},
  number={6},
  pages={240},
  year={1965},
  publisher={APS}
}

@inproceedings{paszke2017automatic,
  title={Automatic differentiation in PyTorch},
  author={Paszke, Adam and Gross, Sam and Chintala, Soumith and Chanan, Gregory and Yang, Edward and DeVito, Zachary and Lin, Zeming and Desmaison, Alban and Antiga, Luca and Lerer, Adam},
  booktitle={NIPS-W},
  year={2017}
}

@book{pointer2019programming,
  title={Programming pytorch for deep learning: Creating and deploying deep learning applications},
  author={Pointer, Ian},
  year={2019},
  publisher={O'Reilly Media}
}

@misc{lippe2024uvadlc,
   title        = {{UvA Deep Learning Tutorials}},
   author       = {Phillip Lippe},
   year         = 2024,
   howpublished = {\url{https://uvadlc-notebooks.readthedocs.io/en/latest/}}
}

@Misc{lightning,
  title = {Torch Lightning}, 
  url = {https://github.com/Lightning-AI/pytorch-lightning}
}

@Misc{Yadan2019Hydra,
  author =       {Omry Yadan},
  title =        {Hydra - A framework for elegantly configuring complex applications},
  howpublished = {Github},
  year =         {2019},
  url =          {https://github.com/facebookresearch/hydra}
}

\end{document}